%% file: iclr2027_conference.tex
\documentclass{article}
\usepackage{iclr2027_conference,times}
\input{math_commands.tex}

\usepackage{graphicx}
\usepackage{booktabs}
\usepackage{colortbl}
\usepackage{multirow}
\usepackage{float}
\usepackage{placeins}
\floatstyle{ruled}
\newfloat{algorithm}{tbp}{loa}
\floatname{algorithm}{Algorithm}
\usepackage[hidelinks]{hyperref}
\usepackage{url}

\title{Video-RSI: Recursive Self-Improvement\\of Video Understanding Agents\\via Harness Evolution}
\author{Bingjun Luo \\ Tsinghua University \And Jialin Guo \\ Harbin Engineering University \And Siqi Li \\ Tsinghua University}

\iclrfinalcopy
\begin{document}
\maketitle
\lhead{}
\renewcommand{\headrulewidth}{0pt}
\begin{abstract}
\input{sections/0_abstract}
\end{abstract}
\input{sections/1_introduction}
\input{sections/2_related_work}
\input{sections/3_method}
\input{sections/4_experiments}
\input{sections/5_conclusion}
\subsection*{AI use statement}

We used generative AI tools to assist with code editing,
figure and table preparation, and reviewing the manuscript
for clarity of expression. The authors reviewed the AI-assisted
content and take responsibility for the final manuscript,
code, and scientific claims.

\bibliography{references}
\bibliographystyle{iclr2027_conference}
\end{document}

%% file: math_commands.tex
\usepackage{amsmath,amsfonts,bm}

\def\eqref#1{equation~\ref{#1}}

\def\1{\bm{1}}

\DeclareMathAlphabet{\mathsfit}{\encodingdefault}{\sfdefault}{m}{sl}
\SetMathAlphabet{\mathsfit}{bold}{\encodingdefault}{\sfdefault}{bx}{n}

%% file: sections/0_abstract.tex
Video understanding agents acquire evidence through an executable harness that controls what they observe and how they use those observations. However, execution traces contain only the evidence acquired by the current harness, leaving competing explanations for failure unresolved and limiting the basis for self-improvement. We introduce Video-RSI, a framework for recursive self-improvement in which a video understanding agent uses its own language model to revise its harness. Through active video investigation, the model revisits the original training videos to test competing failure explanations with additional observations, grounding proposed changes in evidence beyond the existing trace. Cost-aware harness evolution turns these diagnoses into reusable revisions and determines which revisions to retain by considering both answer accuracy and visual cost. Across our evaluation settings on video understanding benchmarks, the evolved agent improves accuracy while processing fewer frames and achieves competitive accuracy-efficiency trade-offs against existing video understanding agents. These results demonstrate the potential for video understanding agents to improve their own evidence acquisition and use through harness evolution. Code is available at \url{https://github.com/bingjunluo/Video-RSI}.

%% file: sections/1_introduction.tex
\section{Introduction}
\label{sec:introduction}

Video understanding agents must decide what to observe before they can decide how to answer.
In long videos, relevant events may be brief, temporally distant, or surrounded by visually similar distractions.
Recent agents address this challenge through iterative retrieval, temporal navigation, and tool-guided observation~\citep{wang2024videoagent,zhang2025dvd,lin2026videoseek}.
At each step, the agent must choose where to look, which details to inspect, and whether the evidence is sufficient to answer.
These decisions are governed by an executable \emph{harness} that controls evidence acquisition, processes observations, and supports subsequent reasoning.
The harness shapes both what the model sees and how it uses that evidence, linking answer accuracy to the cost of visual observation.
Improving this executable layer is therefore an opportunity to strengthen video understanding without changing model weights.

\begin{figure}[t]
    \centering
    \includegraphics[width=\linewidth]{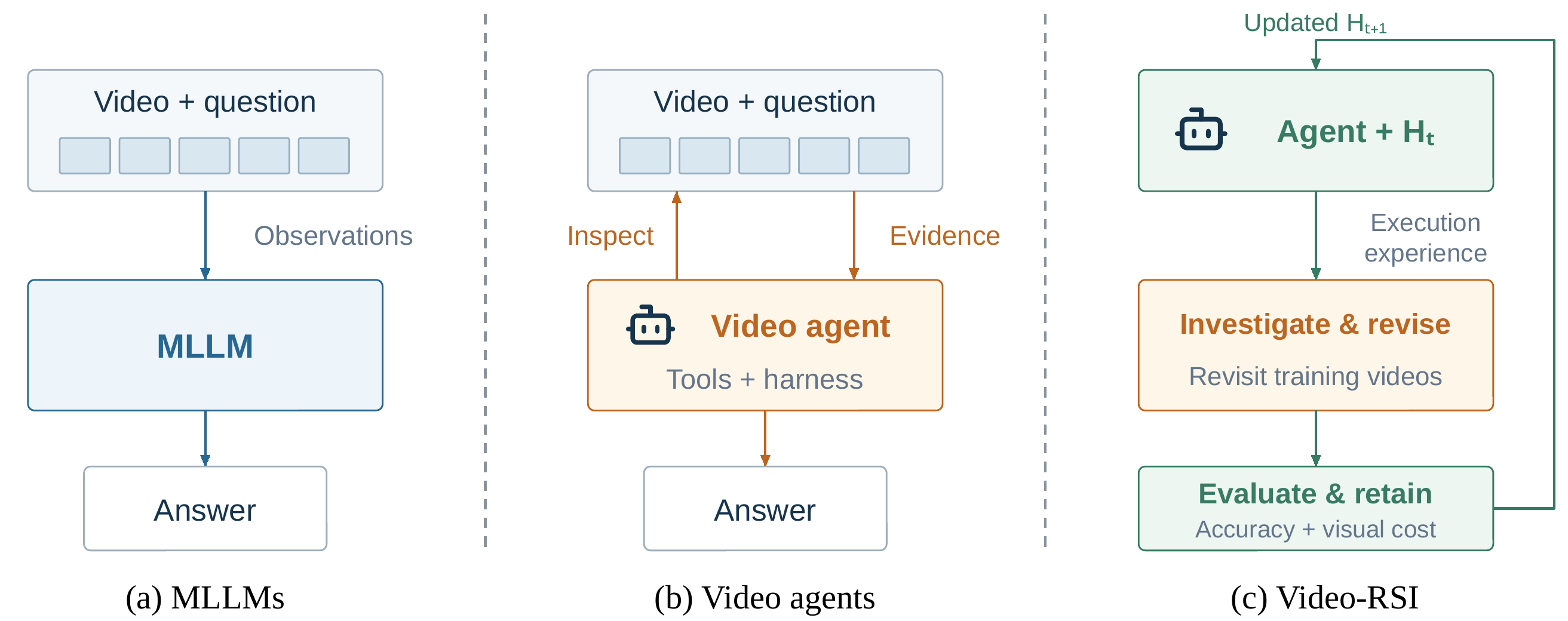}
    \caption{From answering video questions to improving the harness. (a) Direct MLLM inference uses supplied observations. (b) Video agents adapt evidence acquisition within a question. (c) Video-RSI uses the same frozen language model to investigate execution failures, revisit training videos, and revise the harness offline. Revisions are selected by accuracy and visual cost, producing a reusable harness for subsequent questions.}
    \label{fig:intro}
\end{figure}

Automated agent design uses feedback to improve agent programs~\citep{hu2024adas,lee2026metaharness}, and recent work extends code-level evolution to video understanding~\citep{xu2026videoharness}.
Self-Harness further shows that a frozen model can revise its own operating harness in terminal environments~\citep{zhang2026selfharness}.
We study this form of \emph{harness self-improvement} for video understanding. The language model that answers video questions also investigates failures and revises the code governing its behavior.
As shown in Figure~\ref{fig:intro}, adaptive evidence gathering changes what an agent observes within a question, while harness self-improvement changes the procedure it uses across subsequent questions.
This makes execution experience a basis for revising how evidence is acquired and organized, allowing useful changes to accumulate in the program.
Our objective is to turn execution experience into reusable improvements that increase answer accuracy while reducing visual observation, with the underlying models fixed.

This objective requires both evidence for a revision and a criterion for deciding whether to retain it.
An execution trace contains only the observations acquired by the current harness.
A missing event may therefore reflect a sampling gap, a perceptual error, or a failure to use available evidence, each requiring a different modification.
For instance, an event absent from the recorded observations does not establish whether the agent failed to inspect it or misinterpreted what it saw.
Revisiting the video can help distinguish these explanations before the code is changed.
Yet correcting a failure does not necessarily produce a better harness. Denser sampling may improve an answer while increasing visual cost across many questions, whereas better evidence reuse may avoid redundant inspection.
Useful self-improvement must connect a supported failure diagnosis to an assessment of the revised harness's accuracy and visual cost over complete executions.

We introduce \textbf{Video-RSI}, a framework in which the same frozen language model serves as both Solver and Editor to recursively improve a video agent's executable harness.
\emph{Active Video Investigation} revisits training videos to test failure explanations and inform code revisions.
\emph{Cost-Aware Harness Evolution} evaluates the resulting candidates by answer accuracy and visual cost, retaining revisions that become the starting point for further improvement.
These mechanisms link evidence for what to change with criteria for which changes should accumulate.
The resulting loop can revise observation strategies and evidence-processing routines together, assessing their combined effect on the behavior of the complete agent.
Evolution takes place offline. The selected harness is then frozen for evaluation on new questions.
Across four video understanding benchmarks, the evolved harness improves accuracy while processing fewer frames than its initial version and achieves competitive accuracy--efficiency trade-offs against existing video agents.
On MLVU, active investigation also yields higher final accuracy than trace-only revision at similar inference-time frame usage.

Our contributions are threefold:
\begin{itemize}
    \item We introduce Video-RSI, a framework for video harness self-improvement in which the same language model solves video questions, investigates failures, and revises its executable harness while model weights remain fixed.
    \item We develop an improvement loop that links active video investigation to diagnosis-guided code revision and accuracy--cost selection, addressing both the evidence for a modification and its value for subsequent executions.
    \item We evaluate Video-RSI across four video understanding benchmarks and analyze investigation and evolution dynamics, showing improvements over the initial harness in both accuracy and frame usage and the benefit of investigation over trace-only revision in the MLVU comparison.
\end{itemize}

%% file: sections/2_related_work.tex
\section{Related Work}
\label{sec:related_work}

\subsection{Agentic Video Understanding}

Video agents address long videos by deciding which evidence to acquire and how to organize it for reasoning.
Iterative retrieval and hierarchical representations allow VideoAgent and VideoTree to concentrate on question-relevant content~\citep{wang2024videoagent,wang2024videotree}, while VCA guides exploration with intrinsic rewards and a bounded visual memory~\citep{yang2024vca}.
Beyond frame selection, document retrieval~\citep{ma2024drvideo} and multi-granular tool use~\citep{zhang2025dvd,lin2026videoseek} connect video observations to successive reasoning steps.
DVD combines global browsing, clip search, and direct frame inspection. VideoSeek provides complementary overview, skim, and focus operations.
MR.Video uses MapReduce to reconcile entities and aggregate question-specific analyses across segments~\citep{pang2025mrvideo}, while LVAgent combines the observations and reasoning of multiple agents through dynamic collaboration~\citep{chen2025lvagent}.

Another line of work learns observation policies through model training.
FrameThinker uses supervised fine-tuning and reinforcement learning to coordinate reasoning with additional frame acquisition~\citep{he2025framethinker}.
EVA further learns to allocate temporal windows, frame counts, and spatial resolution, and uses failure cases to guide the generation of additional training questions~\citep{zhang2026eva}.
Video-RSI instead keeps the underlying models fixed and evolves the executable harness that governs evidence acquisition and use.
Its improvement loop revisits training videos to diagnose failures before making program changes that apply across questions.

\subsection{Automated Agent Design and Self-Improvement}

Automated agent design treats the programs surrounding language models as an optimization space.
ADAS searches over agents expressed in code, while AFlow searches executable workflows~\citep{hu2024adas,zhang2024aflow}.
These formulations allow changes to the composition and control of model calls beyond isolated prompt edits.
Meta-Harness extends this direction by using historical code, evaluation scores, and execution traces to optimize harnesses around a fixed model~\citep{lee2026metaharness}.
This program-level view motivates our choice of an executable harness as the object of improvement, encompassing both evidence-processing tools and the logic that uses their outputs.

Self-improvement also concerns which system proposes the changes.
STOP studies an improvement program that modifies itself~\citep{zelikman2023stop}, and the Darwin G\"odel Machine combines self-modification with exploration over an archive of agents~\citep{zhang2025dgm}.
Self-Harness uses the same frozen model to propose harness revisions and validates candidates through regression testing~\citep{zhang2026selfharness}.
Video-RSI builds on this code-level self-improvement setting, with the same language model serving as Solver and Editor.
Our focus is the evidence available to that Editor. An execution trace records only what the current harness observed, so investigating the original video can distinguish failure explanations that would otherwise motivate different revisions.

\subsection{Self-Evolving Video Agents}

Video self-improvement has been explored through the generation and refinement of training signals.
EvoGround develops a proposer--solver learning loop for temporal grounding, Video-Zero organizes questioner--solver co-evolution around local temporal evidence, and EvoVid emphasizes temporal information in self-evolution~\citep{jung2026evoground,zhang2026videozero,huang2026evovid}.
These methods improve video capabilities through model training.
Video-RSI instead accumulates improvements in executable code while keeping model weights fixed.

VideoHarness-RSI is a closely related approach to video harness evolution~\citep{xu2026videoharness}.
It evolves context-construction programs that can coordinate retrieval, evidence representations, and auxiliary model calls around frozen vision--language models, and promotes candidates that improve validation accuracy.
Video-RSI shares this program-level search setting and focuses on how the Editor obtains evidence for a revision. Active investigation of the original videos tests failure explanations before code is edited.
Our selection rule also accounts for visual cost, admitting accuracy gains with bounded cost growth or cost reductions with bounded accuracy loss.
Together, investigation and selection connect reusable harness changes to their effects on answer quality and observation requirements.

%% file: sections/3_method.tex
\section{Video-RSI}
\label{sec:method}

\begin{figure}[t]
    \centering
    \includegraphics[width=\linewidth]{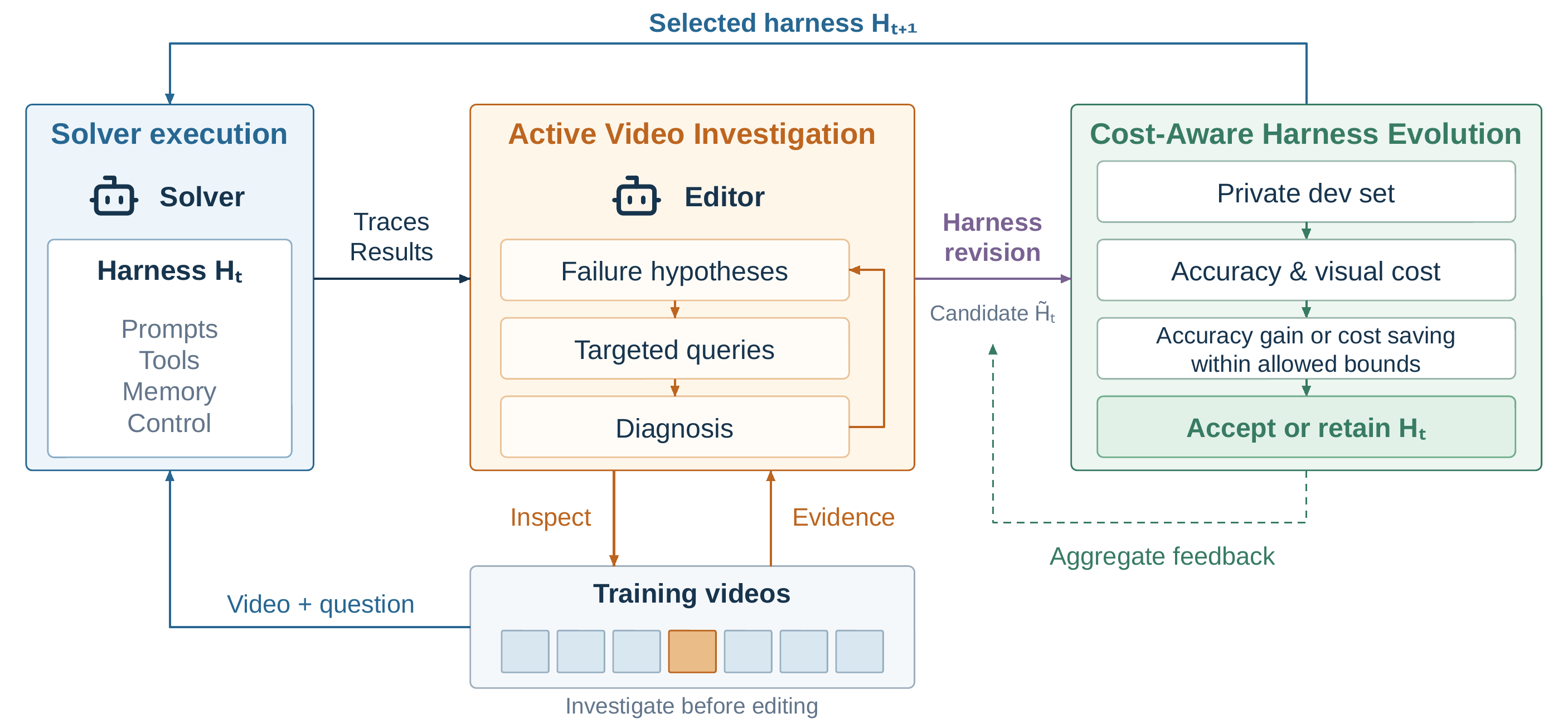}
    \caption{Overview of Video-RSI. The Solver and Editor share a frozen language model. Active video investigation revisits training videos to diagnose failures and guide harness revisions. Cost-aware harness evolution selects candidates using accuracy and visual cost on a private development set, carrying accepted revisions into subsequent iterations.}
    \label{fig:framework}
\end{figure}

Video-RSI evolves the executable harness of a video question-answering agent while keeping the underlying models fixed.
An Editor revisits the original video to investigate failures, uses the resulting diagnosis to revise the harness, and submits the candidate to a performance--cost gate.
Accepted revisions become the starting point for the next evolution attempt.
Figure~\ref{fig:framework} provides an overview.

\subsection{Problem Formulation and Evolvable Harness}
\label{sec:method:setup}

A harness $H$ governs how an agent acquires and uses evidence to answer a video question.
Given a video $V$, a question $q$ with answer choices, and fixed model services $\mathcal{M}$, the Solver produces an answer $\hat y$, an execution trace $\tau$, and a resource record $c$:
\begin{equation}
    (\hat y,\tau,c)=\operatorname{Run}(H,V,q;\mathcal{M}).
    \label{eq:execution}
\end{equation}
The trace records tool calls, observations, and decisions.
The Editor can jointly revise the harness code for prompt construction, tool implementation, observation processing, memory, and execution control.

These components connect evidence acquisition to decision-making by determining what to inspect, how observations are represented and retained, and when to gather more evidence or answer.
Their coordination matters because even a correct observation can lose its temporal or entity associations before reaching the answering step.
Harness evolution can therefore change both the observations acquired and how they support reasoning.

We use a training set $\mathcal{D}_{\mathrm{tr}}$ for investigation and revision and a fixed, video-disjoint selection set $\mathcal{D}_{\mathrm{g}}$.
Selection examples and per-item outcomes remain private to the evaluator.

\subsection{Framework Overview}
\label{sec:method:overview}

Video-RSI uses the same frozen language model as Solver and Editor, accumulating improvements in executable code~\citep{xu2026videoharness}.
At each attempt, the Solver executes $H_t$ on training questions. The Editor examines its traces and revisits videos to test failure explanations before proposing a candidate $\widetilde H_t$.
Independent evaluation on the private selection set determines whether the candidate replaces $H_t$ for the next attempt.
Investigation occurs offline. Additional observations guide reusable code changes, while the selected harness answers subsequent questions using their own inputs.

\subsection{Active Video Investigation}
\label{sec:method:investigation}

An execution trace may not reveal whether an incorrect answer stems from missing an event, misperceiving it, or failing to use existing evidence.
Active investigation lets the Editor revisit the original video to distinguish these explanations before revising the harness.

\paragraph{From failure explanations to diagnostic queries.}
The Editor examines training outcomes, traces, and code, prioritizing representative failures that additional evidence could clarify.
Queries vary the temporal region, sampling density, modality, or perceptual question to test competing explanations.
For example, denser inspection can reveal whether a short event fell between sampled frames.
The interface supports video observations, speech transcripts, on-screen text, and reruns of the unmodified incumbent. Perception requests omit the gold answer.

Let $\mathcal{I}_0$ contain the current code, training traces, available labels, and prior feedback.
The Editor selects queries $u_k$ and accumulates returned evidence $o_k$:
\begin{equation}
\begin{aligned}
u_k &= \operatorname{Query}_{\mathrm{Editor}}(\mathcal{I}_k),\\
o_k &= \operatorname{Investigate}(u_k;H_t,\mathcal{D}_{\mathrm{tr}},\mathcal{M}),\\
\mathcal{I}_{k+1} &= \mathcal{I}_k \cup \{(u_k,o_k)\}.
\end{aligned}
\label{eq:investigation}
\end{equation}
Each query specifies a training example and the observation needed to test an explanation.
The Editor interprets the accumulated evidence to choose the next query, while $H_t$ remains unchanged.

\paragraph{From new evidence to a revision hypothesis.}
New observations can support, contradict, or leave an explanation unresolved, redirecting subsequent queries.
Finding a missed event supports revising temporal coverage. If it was already observed, the diagnosis instead examines how its evidence was represented and used.
Investigation ends with a supported revision hypothesis or an unresolved case when no further informative query is available.

The Editor compares relevant cases to identify a reusable change, its applicable conditions, and successful behaviors to preserve.
Training annotations may locate informative intervals during diagnosis, but the revision must use cues available to the Solver on new questions.
The diagnosis therefore connects observed failures to a procedure for finding and using evidence without knowing the answer or event location in advance.

\subsection{Cost-Aware Harness Evolution}
\label{sec:method:revision}

\paragraph{Diagnosis-guided revision.}
The Editor maps the diagnosis to code changes, specifying when they apply and which successful behaviors to preserve.
Revisions can span the harness components in Section~\ref{sec:method:setup}, such as changing inspection cues or preserving temporal relationships in observations and memory.
A reusable change may require coordinated edits to a tool's output and the code that consumes it.
The Editor can debug candidates through training-side trials, distinct from the pre-edit investigation, before submitting a candidate for selection.

\paragraph{Cost-aware selection.}
\label{sec:method:selection}

After execution checks, the evaluator compares the candidate and incumbent on the same private selection examples under matched conditions.
It measures answer accuracy and visual cost, defined as mean frames processed by successful visual-model calls over complete question-answering executions.
This end-to-end assessment matters because a more elaborate tool can reduce later inspection, while a locally useful repair can increase overall visual usage.

Let $A_t,C_t$ denote the incumbent's accuracy and mean frame count on the selection set, and $\widetilde A_t,\widetilde C_t$ the corresponding candidate measurements.
With $\Delta A_t=\widetilde A_t-A_t$, the gate admits two improvement directions:
\begin{equation}
\begin{aligned}
g_t ={}& \underbrace{[\Delta A_t>0]\land
    [\widetilde C_t\leq(1+\alpha)C_t]}_{\text{accuracy gain with bounded cost growth}}\\
&\lor\underbrace{[-\epsilon\leq\Delta A_t\leq0]\land
    [\widetilde C_t\leq(1-\beta)C_t]\land[C_t>0]}_{\text{cost reduction with bounded accuracy loss}}.
\end{aligned}
\label{eq:update}
\end{equation}
Here, $\alpha\geq0$ bounds relative cost growth, $0<\beta<1$ sets the required relative cost reduction, and $\epsilon\geq0$ bounds accuracy loss. Their values are fixed during evolution and specified in the experimental setup.
If $g_t=1$, the candidate becomes $H_{t+1}$. Otherwise, $H_{t+1}=H_t$.
The Editor receives only aggregate accuracy and frame counts to guide subsequent revisions. Per-example gate traces remain private.
The selected harness is finally evaluated on held-out data.
Algorithm~\ref{alg:evolution} summarizes the complete evolution process.

\begin{algorithm}[H]
\caption{Recursive harness evolution in Video-RSI}
\label{alg:evolution}
\small
\textbf{Input:} initial harness $H_0$, fixed services $\mathcal{M}$, training set $\mathcal{D}_{\mathrm{tr}}$, private gate $\mathcal{D}_{\mathrm{g}}$, attempt budget $T$.\\
\textbf{Output:} final accepted harness $H_T$.
\begin{tabbing}
\quad\=\quad\=\quad\=\kill
\textbf{for} $t=0,\ldots,T-1$ \textbf{do}\\
\> Execute $H_t$ on training examples and inspect its traces.\\
\> Initialize investigation context $\mathcal{I}_0$ and set $k=0$.\\
\> \textbf{while} further investigation is warranted and resources permit \textbf{do}\\
\>\> Select a query to distinguish unresolved failure explanations.\\
\>\> Acquire evidence and construct $\mathcal{I}_{k+1}$ using Eq.~(\ref{eq:investigation}).\\
\>\> Revise the supported explanations and remaining questions.\\
\>\> Set $k\leftarrow k+1$.\\
\> \textbf{end while}\\
\> Synthesize a diagnosis and plan a reusable behavior change.\\
\> Edit $H_t$ into a candidate $\widetilde H_t$ and run training-side checks.\\
\> Validate and compare the candidate with $H_t$ on $\mathcal{D}_{\mathrm{g}}$.\\
\> Accept or retain the incumbent according to Eq.~(\ref{eq:update}).\\
\textbf{end for}\\
\textbf{return} $H_T$
\end{tabbing}
\end{algorithm}

%% file: sections/4_experiments.tex
\section{Experiments}
\label{sec:experiments}

\subsection{Experimental Setup}
\label{sec:experiments:setup}

\paragraph{Benchmarks and data splits.}
We evaluate on the long-video splits of Video-MME~\citep{fu2024videomme} and LongVideoBench~\citep{wu2024longvideobench} with subtitles, the official public subset of EgoSchema~\citep{mangalam2023egoschema}, and MLVU test~\citep{zhou2024mlvu}.
For evolution, we use 216 LVBench questions~\citep{wang2025lvbench} for investigation and revision, and another 72 questions as a fixed, video-disjoint development set for candidate selection.
Development examples and per-question outcomes remain private. The Editor receives only aggregate feedback.

\paragraph{Models and evolution protocol.}
The Solver and Editor use DeepSeek-V4-Pro, with Qwen3.6-Plus for visual observations. All model weights are frozen.
The initial harness $H_{S0}$ supports global and local video inspection, transcripts, OCR, and observation memory.
We run 20 revision attempts using Eq.~(\ref{eq:update}) with $\alpha=0.1$, $\beta=0.2$, and $\epsilon=1/|\mathcal{D}_{\mathrm{g}}|$, allowing at most one fewer correct answer in the cost-reduction branch.
The final accepted harness is frozen for the main evaluation, while intermediate accepted versions are evaluated to analyze evolution dynamics.
Active-investigation and trajectory-only variants share training examples, annotation access, initialization, models, and per-attempt resource limits, differing only in pre-edit access to new video observations.

\paragraph{Baselines.}
We compare with MLLMs, video agentic models, and harness self-improvement methods.
We reproduce VideoSeek~\citep{lin2026videoseek} and VideoHarness-RSI~\citep{xu2026videoharness} using the same model (DeepSeek-V4-Pro) as Video-RSI. Results for the remaining methods are taken from their original publications.
The initial harness $H_{S0}$ serves as the reference for evolution gains. Trajectory-only revision and an accuracy-only gate assess the contributions of investigation and candidate selection.

\paragraph{Evaluation metrics.}
We report multiple-choice accuracy and mean visual frames per question.
For our evaluations, frame usage sums all frames processed by successful visual-model calls during each complete execution, then averages across questions. Offline Editor observations are excluded.

\subsection{Main Results}
\label{sec:experiments:main}

Table~\ref{tab:main_results} places Video-RSI alongside MLLMs, video agentic models, and harness self-improvement methods.
Our analysis focuses on the reproduced VideoSeek and VideoHarness-RSI implementations. Published results provide broader context.
With its harness evolved on LVBench and frozen for evaluation, Video-RSI achieves higher accuracy than both reproduced baselines across all four benchmarks, with lower frame usage in all but the VideoSeek comparison on Video-MME.

\input{tables/main_results}

Relative to VideoSeek, the accuracy--cost balance varies across benchmarks.
On MLVU, Video-RSI gains 4.8 percentage points while using 27.2\% fewer frames, combining better answers with reduced visual usage.
On LongVideoBench, the main benefit is substantially fewer frames at similar accuracy. On EgoSchema, both accuracy and frame usage improve.
On Video-MME, higher accuracy accompanies a moderate increase in frame usage.
Thus, the advantage is not a uniform reduction in observation, but a favorable balance between answer quality and visual usage across different evaluation settings.

Compared with VideoHarness-RSI, Video-RSI achieves higher accuracy with fewer frames on every benchmark.
The accuracy gain is largest on Video-MME, whereas EgoSchema shows closely comparable accuracy with substantially lower frame usage.
The evolved harness thus offers two practical benefits by improving answer quality and reducing the observations needed for comparable performance.
The controlled comparisons in Section~\ref{sec:experiments:investigation} further examine how active investigation and cost-aware selection contribute to the resulting harness.

\subsection{Ablation Study}
\label{sec:experiments:investigation}

Table~\ref{tab:active_investigation} examines how harness evolution depends on the evidence available before revision and the criterion used to retain candidates.
The trajectory-only variant restricts the Editor to existing execution traces, while active investigation allows additional video observations before editing. Both use 20 revision attempts under the shared conditions in Section~\ref{sec:experiments:setup}.
The accuracy-only gate variant examines selection without the visual-cost criterion.
We evaluate the resulting harnesses on all four benchmarks, using the initial harness as a reference.

\input{tables/active_investigation}

\paragraph{Active investigation improves the resulting harness.}
Active investigation yields higher final accuracy than trajectory-only revision on every benchmark.
On MLVU, the gain reaches 8.2 percentage points with similar inference-time frame usage. On the other three benchmarks, higher accuracy accompanies fewer frames.
This pattern links additional evidence during offline revision to better deployed behavior without a consistent increase in inference-time observation.
The investigation-enabled run also retains five of twenty revisions, compared with three for trajectory-only revision.
Alongside the endpoint results, these observations support using evidence beyond the existing trace to guide reusable harness changes.

\paragraph{Cost-aware selection improves the accuracy--cost balance.}
The accuracy-only gate improves accuracy over the initial harness on all four benchmarks, but increases frame usage on three.
Video-RSI achieves higher accuracy than this variant while using fewer frames on every benchmark. On Video-MME, it uses roughly half as many frames.
Thus, the cost-aware run obtains visual savings without sacrificing endpoint accuracy in these evaluations.
The comparison shows why accuracy gains alone do not fully characterize a useful harness revision. Selecting for both objectives can retain more efficient behavior while preserving improvements in answer quality.

\subsection{Evolution Dynamics}
\label{sec:experiments:evolution}

\begin{figure}[htbp]
    \centering
    \vspace{-5mm}
    \includegraphics[width=0.85\linewidth]{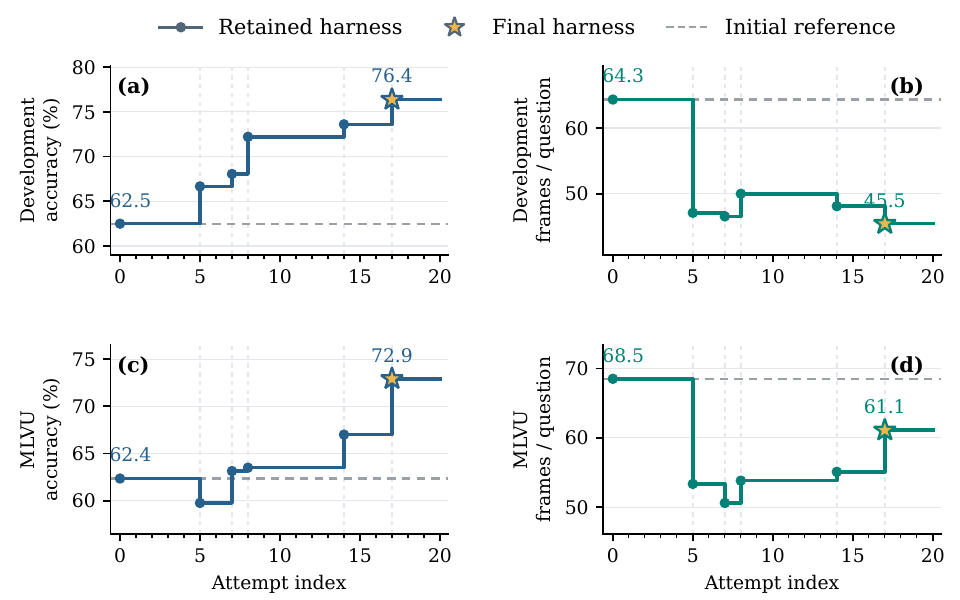}
    \vspace{-2mm}
    \caption{Evolution over 20 revision attempts. Accuracy and mean frames per question on (a,b) the development set and (c,d) MLVU. Curves track the retained harness. Vertical guides mark accepted revisions, and flat intervals indicate an unchanged incumbent.}
    \vspace{-5mm}
    \label{fig:evolution}
\end{figure}

Figure~\ref{fig:evolution} tracks the retained harness over 20 revision attempts on the development set and MLVU.
The MLVU evaluations were performed retrospectively after evolution and did not inform candidate selection.
On the development set, the first accepted revision produces most of the net reduction in frame usage, while subsequent revisions steadily improve accuracy with smaller, nonmonotonic changes in visual cost.
The run thus combines an early efficiency gain with continued refinement of answer quality.

The MLVU trajectory reveals that improvements on the selection set need not transfer at every step.
The first accepted revision reduces frame usage but lowers MLVU accuracy, despite improving both development metrics.
Accuracy recovers with the next revision, and the final two accepted revisions account for most of the net MLVU accuracy gain.
The final harness therefore reflects cumulative refinement beyond the initial efficiency improvement. An early development gain alone would give an incomplete picture of its evaluation behavior.

Accuracy and visual cost also follow different paths during later refinement.
From $H_{S2}$ onward, MLVU accuracy rises together with frame usage, yet the final harness remains more accurate and uses fewer frames than $H_{S0}$.
This pattern is consistent with the selection rule, which permits bounded cost increases for accuracy gains, rather than requiring frame usage to decrease at every update.
Later revisions can use more frames while retaining a net efficiency advantage over the initial harness, improving the final accuracy--cost balance.

\subsection{Analysis of Evolved Harnesses}
\label{sec:experiments:harnesses}

Comparing the initial harness with the retained revisions reveals three changes in how the Solver acquires and uses evidence:

\textbf{Structured perception.} In $H_{S2}$, free-form visual responses give way to goal-directed observations with structured findings, image references, and explicit uncertainty. The harness maps image references to frame timestamps, preserving the link between a finding and its temporal evidence. This interface lets follow-up reasoning target unresolved details and their supporting frames without changing model weights.

\textbf{Evidence reuse through relation graphs.} Introduced in $H_{S4}$, the graph mechanism extracts entities, events, and relations from accumulated visual, transcript, and OCR observations. For eligible relational or temporal questions, it returns evidence gaps that can guide further inspection. Graph queries themselves process no new frames, making evidence organization an intermediate step between existing observations and additional acquisition.

\textbf{Selective observation routines.} The Global-64 route in $H_{S3}$ expands the initial 32-frame overview for selected whole-video questions. The occurrence tool in $H_{S5}$ defines counting units, verifies candidate events, and consolidates overlapping evidence to avoid duplicate counts. These routines coexist with transcript-based localization and observation reuse. The final harness improves accuracy while using 10.8--44.1\% fewer frames than $H_{S0}$ across the four benchmarks, showing that richer observation tools can coexist with lower aggregate visual usage.

These retained changes illustrate how evolution modifies both evidence acquisition and its use in reasoning. The reported savings measure visual frames rather than total model computation.

%% file: tables/main_results.tex
\begin{table}[t]
\centering
\caption{Video understanding results. Acc. denotes accuracy (\%) and Frames denotes visual frame usage. Bold indicates column-best values. $\dagger$ denotes official M-Avg.}
\label{tab:main_results}
\small
\setlength{\tabcolsep}{3pt}
\renewcommand{\arraystretch}{1.08}

\begin{tabular}{@{}lrrrrrrrr@{}}
\toprule
\multirow{3}{*}{Method}
& \multicolumn{2}{c}{MLVU}
& \multicolumn{2}{c}{LongVideoBench}
& \multicolumn{2}{c}{Video-MME}
& \multicolumn{2}{c}{\multirow{2}{*}{EgoSchema}} \\
& \multicolumn{2}{c}{Test}
& \multicolumn{2}{c}{Long val}
& \multicolumn{2}{c}{Long}
& \multicolumn{2}{c}{} \\
\cmidrule(lr){2-3}
\cmidrule(lr){4-5}
\cmidrule(lr){6-7}
\cmidrule(l){8-9}
& Acc.$\uparrow$ & Frames$\downarrow$
& Acc.$\uparrow$ & Frames$\downarrow$
& Acc.$\uparrow$ & Frames$\downarrow$
& Acc.$\uparrow$ & Frames$\downarrow$ \\

\midrule
\rowcolor[gray]{0.92}
\multicolumn{9}{l}{\textit{MLLMs}} \\
\midrule
GPT-4o
& 54.9$^{\dagger}$ & 0.5 fps
& 60.9 & 256
& 65.3 & 384
& 70.4 & -- \\
Gemini 1.5 Pro
& -- & --
& 58.6 & 256
& 67.4 & 1233
& -- & -- \\
Gemini 2.0 Flash
& -- & --
& 45.7 & 256
& 63.0 & 1233
& 71.2 & -- \\
Qwen2.5-VL-72B
& -- & --
& -- & --
& 53.2 & 256
& -- & -- \\
GPT-5
& -- & --
& 64.5 & 384
& 67.9 & 384
& -- & -- \\

\midrule
\rowcolor[gray]{0.92}
\multicolumn{9}{l}{\textit{Video Agentic Models}} \\
\midrule
VideoAgent
& -- & --
& -- & --
& 46.4 & 24.6
& 60.2 & 8.4 \\
VideoTree
& -- & --
& -- & --
& 53.1 & 98.0
& 66.2 & 62.4 \\
DrVideo
& -- & --
& -- & --
& 51.7 & 493.2
& 66.4 & -- \\
VCA
& -- & --
& -- & --
& 54.2 & 18.1
& 73.6 & \textbf{7.2} \\
MR.Video
& -- & --
& 61.6 & --
& 63.4 & --
& 73.8 & -- \\
DVD
& -- & --
& 68.6 & 2816
& 67.3 & 4932
& 76.6 & -- \\
FrameThinker
& -- & --
& -- & --
& 47.6 & 24.1
& -- & -- \\
LVAgent
& -- & --
& -- & --
& 74.3 & --
& -- & -- \\
EVA-GRPO
& -- & --
& -- & --
& 48.4 & 22.8
& -- & -- \\
VideoSeek
& 68.1 & 83.9
& 70.9 & 71.9
& 78.7 & \textbf{17.8}
& 73.2 & 72.3 \\

\midrule
\rowcolor[gray]{0.92}
\multicolumn{9}{l}{\textit{Harness Self-Improvement}} \\
\midrule
VideoHarness-RSI
& 64.1 & 83.2
& 68.4 & 92.1
& 69.7 & 92.8
& 78.0 & 66.8 \\
\rowcolor[rgb]{0.90,0.94,0.98}
\textbf{Video-RSI (Ours)}
& \textbf{72.9} & \textbf{61.1}
& \textbf{71.1} & \textbf{41.1}
& \textbf{80.0} & 21.6
& \textbf{78.2} & 42.4 \\
\bottomrule
\end{tabular}

\par\vspace{3pt}
\begin{minipage}{\linewidth}
\footnotesize
\end{minipage}

\end{table}

%% file: tables/active_investigation.tex
\begin{table}[t]
\centering
\caption{Ablations of active investigation and candidate selection. Acc. is accuracy (\%). Frames is mean visual frames per question.}
\label{tab:active_investigation}
\small
\setlength{\tabcolsep}{2.5pt}
\renewcommand{\arraystretch}{1.08}
\begin{tabular}{@{}lrrrrrrrr@{}}
\toprule
\multirow{3}{*}{Variant} & \multicolumn{2}{c}{MLVU} & \multicolumn{2}{c}{LongVideoBench} & \multicolumn{2}{c}{Video-MME} & \multicolumn{2}{c}{\multirow{2}{*}{EgoSchema}} \\
& \multicolumn{2}{c}{Test} & \multicolumn{2}{c}{Long val} & \multicolumn{2}{c}{Long} & \multicolumn{2}{c}{} \\
\cmidrule(lr){2-3}\cmidrule(lr){4-5}\cmidrule(lr){6-7}\cmidrule(l){8-9}
& Acc.$\uparrow$ & Frames$\downarrow$ & Acc.$\uparrow$ & Frames$\downarrow$ & Acc.$\uparrow$ & Frames$\downarrow$ & Acc.$\uparrow$ & Frames$\downarrow$ \\
\midrule
Initial harness & 62.4 & 68.5 & 67.7 & 51.6 & 73.4 & 38.6 & 71.4 & 66.0 \\
\midrule
Trajectories only & 64.7 & \textbf{60.3} & 67.2 & 43.1 & 73.7 & 28.5 & 76.6 & 51.8 \\
Accuracy-only gate & 67.1 & 81.9 & 69.5 & 66.4 & 74.9 & 43.1 & 77.4 & 60.5 \\
Video-RSI & \textbf{72.9} & 61.1 & \textbf{71.1} & \textbf{41.1} & \textbf{80.0} & \textbf{21.6} & \textbf{78.2} & \textbf{42.4} \\
\bottomrule
\end{tabular}
\end{table}

%% file: sections/5_conclusion.tex
\section{Conclusion}
\label{sec:conclusion}

We presented Video-RSI, a framework for improving video understanding agents through harness evolution with fixed model weights.
The same language model serves as Solver and Editor, revisiting training videos to test failure explanations and translating the resulting diagnoses into reusable code changes.
Cost-aware selection retains revisions according to both answer accuracy and visual usage.
Across four benchmarks, the evolved harness improves accuracy while using fewer frames than its initial version, with ablations supporting the contributions of active investigation and cost-aware selection.
Analysis of the retained code reveals changes in structured perception, evidence reuse, and selective observation.
Together, these findings support harness evolution as a practical route to improving how frozen-model video agents acquire and use evidence.